\documentclass[letterpaper, 10 pt, twoside]{IEEEtran} 

\usepackage{amsmath,amsfonts}
\usepackage{algorithm}
\usepackage{array}
\usepackage[caption=false,font=normalsize,labelfont=sf,textfont=sf]{subfig}
\usepackage{textcomp}
\usepackage{stfloats}
\usepackage{url}
\usepackage{verbatim}
\usepackage{graphicx}
\usepackage{booktabs}
\usepackage{cite}
\usepackage{float}
\usepackage{diagbox} 
\usepackage{multirow} 
\usepackage{makecell}
\usepackage{enumitem}
\usepackage{algpseudocode}
\usepackage{xcolor} 
\usepackage{siunitx}
\usepackage{tikz}
\title{\LARGE \bf
SlipSense: Multimodal Sensing for Online Slip Detection in Legged Robots
}

\author{Iris Szu-Yao Liu$^{1,2}$, Chien Chern Cheah$^{1}$, Meng Yee (Michael) Chuah$^{2}$%
\thanks{This research is supported by the Home Team Science and Technology Agency (HTX), and the A*STAR RIE2030 CDF - H26-KSR0065.}%
\thanks{$^{1}$ School of Electrical and Electronic Engineering, Nanyang Technological University, Singapore}%
\thanks{$^{2}$Institute for Infocomm Research, Agency for Science Technology and Research, Singapore} %
}

\newcommand\copyrighttext{%
  \footnotesize \textcopyright 2026 IEEE.  Personal use of this material is permitted. Permission from IEEE must be obtained for all other uses, in any current or future media, including reprinting/republishing this material for advertising or promotional purposes, creating new collective works, for resale or redistribution to servers or lists, or reuse of any copyrighted component of this work in other works.}
\newcommand\copyrightnotice{%
\begin{tikzpicture}[remember picture,overlay]
\node[anchor=south,yshift=4pt] at (current page.south) {\parbox{\dimexpr\textwidth-\fboxsep-\fboxrule\relax}{\copyrighttext}};
\end{tikzpicture}%
}

\begin{document}
\maketitle
\copyrightnotice

\markboth{IEEE INTERNATIONAL CONFERENCE ON ROBOTICS AND AUTOMATION (ICRA). PREPRINT VERSION. ACCEPTED JANUARY 2026}%
{LIU \MakeLowercase{\textit{et al.}}: SLIPSENSE: MULTIMODAL SENSING FOR ONLINE SLIP DETECTION IN LEGGED ROBOTS}

\thispagestyle{empty}
\pagestyle{empty}

\begin{abstract}
Legged robots rely on accurate ground interaction awareness to traverse variable terrains, such as slippery surfaces. Existing slip detection methods often rely on kinematics and proprioception, which lack the sensitivity to detect early-stage slips that occur prior to catastrophic instability. Thus, this paper presents SlipSense, a novel framework for online force-based slip detection using a custom lightweight sensorized foot for quadrupeds to detect slip. The framework integrates a multimodal sensor design with a LSTM-based model to infer ground reaction forces and detect slip-indicative anomalies during locomotion. The proposed framework is deployed on a Unitree Go1 quadruped to demonstrate blind online slip detection over a slippery terrain. Our method detects early-stage slips down to an average displacement of 24.1 ± 6.4mm with an overall accuracy of 85.9\%. This represents a 3.3-fold finer detection resolution and a 24\% relative accuracy improvement over a standard kinematic baseline that uses foot velocity inferred through state estimation. The work in this paper serves as a foundation for force-aware gait adaptation in legged robotic locomotion, allowing future controllers to estimate terrain friction and adjust constraints, thus improving the overall stability of the system.

\end{abstract}

\section{Introduction}
Recent years have seen an increasing number of legged robots deployed for navigating complex environments such as for exploration or industrial automation.  These systems are often subject to highly dynamic and unpredictable environments that may involve terrains with varying friction. This remains a challenge to achieve safe and efficient locomotion in versatile terrains due to the risk of slip and instability. In such cases, multidimensional ground reaction forces (GRFs) are a crucial source of information for robust legged robot control, as it provides direct insights into the robot's interaction with the environment. Conventional approaches use a combination of proprioception and state estimation to perform GRF estimation \cite{bledt2018contact}\cite{bosworth2015super}. However, these methods often rely on precise modeling of leg mechanics and are highly sensitive to unmodeled inertial effects or joint friction. Consequently, slip detection strategies built on this principle, such as those that infer slip from foot velocity \cite{focchi2018slip}\cite{yan2024slip}, must wait for a significant slip to occur before its signal can be distinguished from the noise of normal locomotion. This leads to a compromise of setting a high detection threshold to prevent false positives at the cost of insensitivity to early-stage slips that precede unrecoverable stability. Alternatively, vision-based terrain mapping has also been explored \cite{fankhauser2018robust}, but it relies on accurate elevation maps and fails under occlusion or poor visibility, which can severely degrade stability. Therefore, it is desirable to integrate a direct force sensing system into legged robots to perform real-time GRF estimation. However, practical integration of foot-mounted sensors face challenges such as durability, bulkiness, and the ability to operate under dynamic locomotion. These hardware challenges have limited the adoption of sensorized feet into dynamic locomotion applications, despite their potential for enhancing stability. 

\begin{figure}[t]
    \centering
    \includegraphics[width=1.0\linewidth]{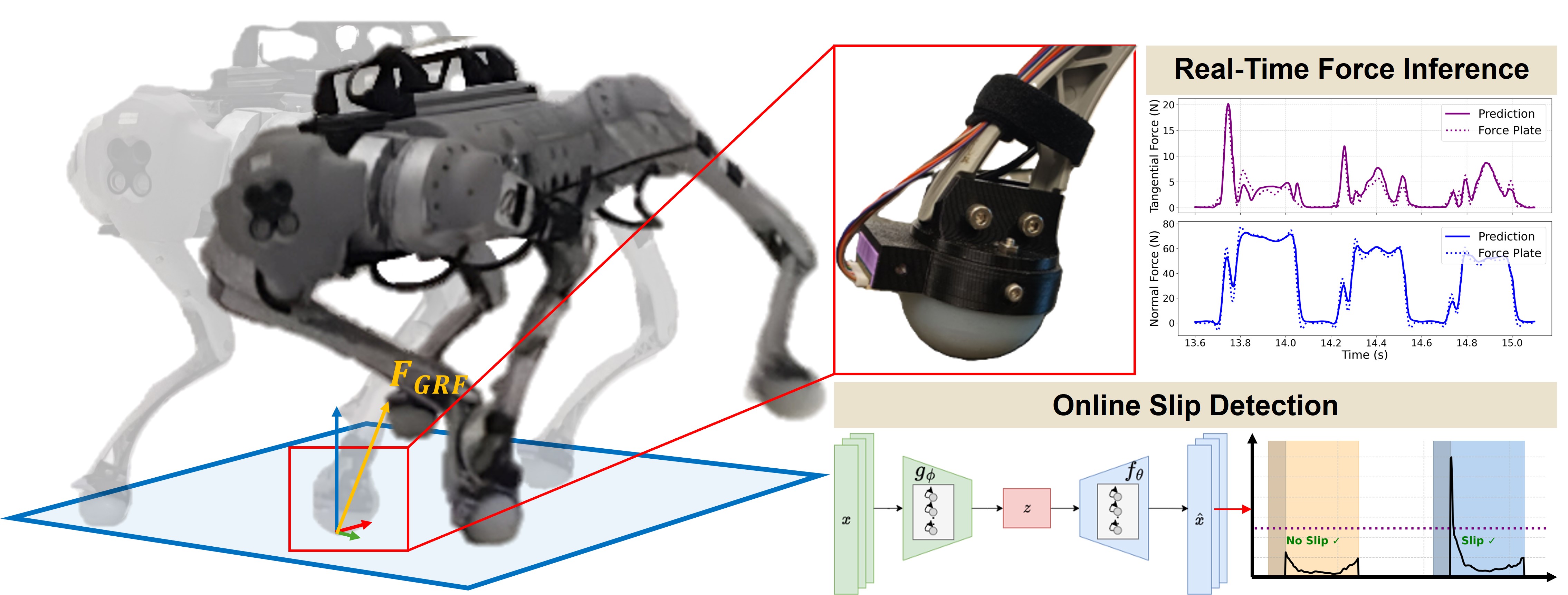}
    \caption{SlipSense: multimodal online slip detection for legged robots using a custom foot-mounted force sensor. The hardware is integrated with dual-temporal learning models to perform real-time force inference and slip-indicative anomaly detection. Figure shows snapshots of robot becoming unstable after slip.}
    \label{fig:cover}
\end{figure}

To address these limitations, we propose a framework which integrates a lightweight, multimodal force sensor with a learning model to provide real-time force inference and high-sensitivity slip detection. To the best of our knowledge, this is the first work to present a deployed multimodal force-based slip detection framework under dynamic quadruped gait, which can serve as a foundation for future force-aware locomotion control. Our contributions are summarized as follows:
\begin{itemize}
    \item \textbf{SlipSense (Fig. \ref{fig:cover}), a complete framework for high-sensitivity slip detection:} we introduce an end-to-end system that integrates a custom multimodal force sensor with a LSTM-based learning model, enabling the detection of early slip events that cannot be detected by conventional kinematic-only methods. 
    \item \textbf{A data-efficient anomaly detection approach for slip detection training:} we frame the slip detection as a one-class self-supervised anomaly detection problem, mitigating the challenge of acquiring accurate, well-labeled slip data for training our model. 
    \item \textbf{Experimental validation on a real quadruped demonstrates a 3.3-fold improvement in detection resolution:} we deployed our framework for blind robot traversal over terrains of varying friction and achieved a higher accuracy and ability to detect smaller slips compared to a kinematic baseline using inferred foot velocity through state estimation. This was validated through a motion capture system.
\end{itemize}

\section{Related Work}
\begin{figure*}[t]
    \centering
    \includegraphics[width=1.0\textwidth]{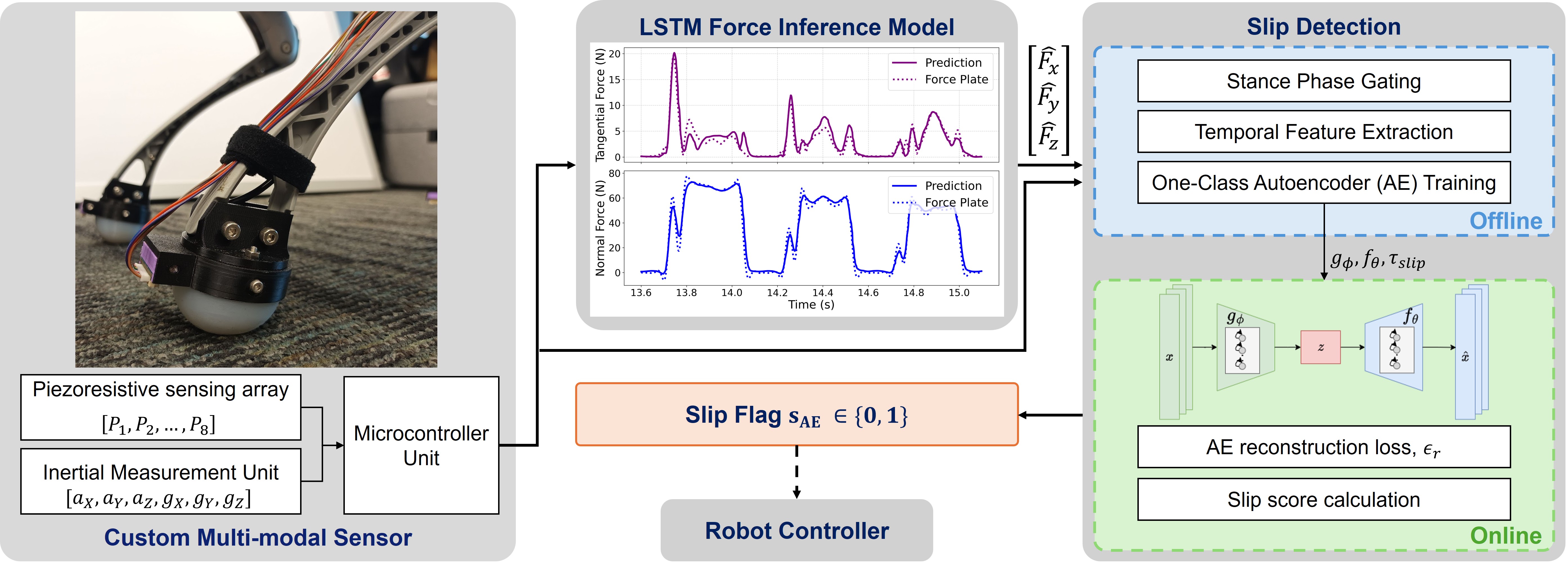}
    \caption{Overview of the proposed SlipSense framework for online slip detection in quadruped robots. The system consists of three modules: (1) a lightweight multimodal footpad sensor combining pressure and inertial measurements, (2) a temporal LSTM-based model that reconstructs 3D GRFs in real-time, and (3) a self-supervised anomaly detection module that fuses kinetic and kinematic features to identify slip events. Integration with the robot’s controller is for future work.}
    \label{fig:framework}
\end{figure*}

Ground reaction force (GRF) sensing plays a critical role in legged robot control, providing low-level feedback for balance, adaptation, and interaction with diverse terrains. Legged robotic force sensing methods fall under two categories, namely state estimation-based and direct force sensing. State estimation-based sensing uses sensorless approaches such as inference from joint torques \cite{bledt2018contact}\cite{bosworth2015super}, extended state observers \cite{chan2013extended}, and generalized momentum observers \cite{cong2020contact}. These approaches often struggle to detect fine sensor-surface contact changes \cite{stone2020walking} and are commonly contaminated by noise due to electromagnetic interference. Direct force sensing methods adopted in studies include piezoresistive pressure \cite{epstein2020bi}\cite{ruppert2020foottile}, capacitive \cite{wu2016integrated}\cite{wu2019tactile}, optical \cite{stone2020walking} \cite{song2024tactid}, and strain-gauge based force torque (F/T) sensors \cite{valsecchi2020quadrupedal}. Sensorized feet in legged robots offer potential in enabling tasks such as terrain characterization \cite{wu2016integrated}\cite{guo2020soft}, friction estimation \cite{song2024tactid}, slip detection \cite{okatani2019mems}\cite{park2019design}, and edge detection \cite{van2024edge}\cite{wiedebach2016edge2}. The focus of this paper will be on legged robot slip detection. Slip detection is essential for maintaining stability during locomotion, especially on low-friction or uncertain terrains and often fall under two categories: direct force-based and kinematic-based methods. 

Force-based approaches rely on additive sensors integrated into the foot which directly estimates the ground contact forces. An example is presented in \cite{okatani2019mems} where a MEMS slip sensor combined with a force sensor and shock absorber is used to measure slipperiness. However, this design is limited to vertical forces up to only 20N and introduced significant inertia due to bulkiness. Alternatively, Park et al. \cite{park2019design} proposed an anti-skid foot design that detects slip when excessive tangential force is applied to the foot, triggering the release of anchoring spines to prevent slip. Slip detection was demonstrated in \cite{epstein2020bi} using a piezoresistive stress field sensor on a detached Mini Cheetah leg in a quasi-static setup. More generally, previous studies highlight the fragility and limited durability of additive foot force sensors as a key bottleneck for practical force-based slip detection \cite{focchi2018slip}\cite{nistico2022slip}. Thus, while force-based methods offer greater sensitivity, no existing design has demonstrated the combination of a lightweight, high force-range, multi-axis sensor reliable for deployment on a dynamic quadruped for slip detection. These works are often limited in practical integration due to two key challenges: fragility under high-impact and adverse inertial effects from bulky designs.

Alternatively, kinematic-based slip detection methods rely on proprioceptive information, where slippage is inferred from foot motion, typically using joint encoders and IMUs \cite{bloesch2013state}\cite{teng2021legged}. Commonly, these methods adopt a velocity-level slip detection, such as in \cite{focchi2018slip}, where single and multiple leg slippage are detected by estimating the foot velocity in the body and world frame, respectively. Similarly in \cite{nistico2022slip}, slip is detected when the norm of the difference between the desired and actual foot velocities in the body frame exceeds a certain threshold. Yan et al. \cite{yan2024slip} fused the foot displacement and velocity to estimate the slip state. As another approach, the authors in \cite{maravgakis2023probabilistic} mounted an IMU at the foot to probabilistically determine the contact state stability. In \cite{sun2023learning}, a deep learning approach was taken and a feature set comprising proprioceptive information was used to predict slips in simulation. A common limitation across all kinematic methods is the reliance on indirect, model-dependent data. Error accumulation from IMU integration and state estimation drift results in insensitivity to the early-stage slips that precede larger, unrecoverable slips.

In summary, the literature reveals a critical trade-off in legged robot slip detection methods: force-based methods offer high sensitivity but are fragile and impractical for dynamic deployment, while kinematic methods allow for easier integration but are insensitive to early-stage slip events. This presents a research gap for a system that can achieve both high-fidelity detection and demonstrated robust real-world deployment. To address this, we propose our end-to-end multimodal slip detection framework, named SlipSense, which fuses force and inertial information and leverages them to reliably detect slip, enabling finer sensitivity and more robust online deployment.

\section{Integrated Force Inference and Slip Detection Framework}
The overview of our proposed framework, SlipSense, is illustrated in Fig. \ref{fig:framework}. The pipeline begins with our custom multimodal footpad sensor which outputs raw piezoresistive pressure and inertial measurements. A temporal force inference model processes the sequence of piezoresistive measurements to reconstruct the 3D GRFs, $[\hat{F_X},\hat{F_Y},\hat{F_Z}]$. Coupled with the inertial measurements, the feature set is fed into the slip detection module, which leverages a one-class anomaly detection model to distinguish between stable contacts and anomalous slip behavior.

\section{Sensorized Foot and Force Inference Model}
In the following section, we introduce our custom-made force sensor design and the temporal force inference model for predicting GRFs in real-time. This serves as the initial stage of the pipeline for the downstream slip detection task.
\subsection{Force Sensor Design}
\begin{figure}
    \centering
    \includegraphics[width=1.0\columnwidth]{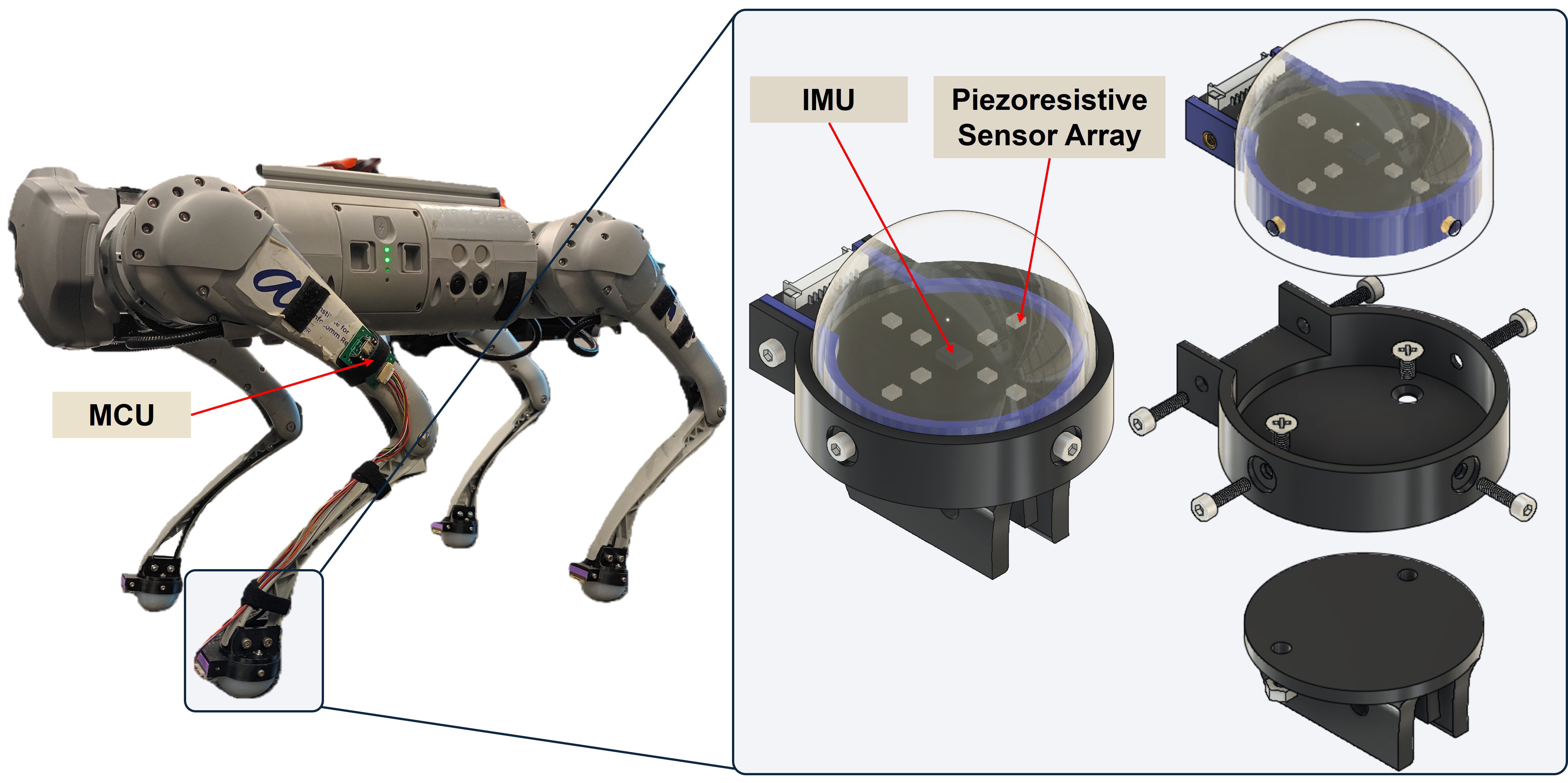}
    \caption{Our custom multimodal footpad sensor mounted onto a Unitree Go1 quadruped. The exploded diagram reveals the modularity of the footpad mounting design and the internal PCB components.}
    \label{fig:sensor}
\end{figure}
Our force sensor (shown in Fig. \ref{fig:sensor}) is based on the stress field sensing concept presented in \cite{epstein2020bi}\cite{chuah2019bi}, which uses an array of piezoresistive sensors embedded within a dome of rubber to reconstruct the 3D forces from a load. However, to enable robust slip detection during dynamic locomotion, we introduced two key modifications:
\begin{itemize}
    \item IMU Integration: An IMU chip is placed at the center of the custom PCB board, surrounded by an array of eight piezoresistive pressure sensors arranged in two concentric diamonds. This arrangement ensures that kinematic and kinetic data are captured from the same rigid body.
    \item Enhanced Durability: Similar to in \cite{chuah2014enabling}, we remove the piezoresistive sensor caps to expose the diaphragm, thus increasing the sensitivity. We automated this process using a 5-axis CNC machine to precisely mill off the sensor caps. To ensure the exposed diaphragm is protected from repeated loads, we pot the sensor cavities with a harder Shore 40A silicone before overmolding with a softer Shore 30A elastomer. The fabricated sensor has a diameter of 45mm and weighs 65g, which is an insignificant increase compared to the original Unitree Go1 foot weighing 51g.
\end{itemize}

\subsection{Force Inference Model}
To map the raw sensor array readings to contact forces, a learning-based approach was chosen to generalize across the various loading conditions of dynamic locomotion. To model the material viscoelastic behavior introduced by the elastomeric material, we use a Long Short Term Memory (LSTM) model to use historical deformation information to predict the current state of the output. Previous works using static models such as Gaussian Process Regression or ANN \cite{epstein2020bi}, often struggle to predict forces at the upper range, due to the more complex material behavior such as hysteresis and stress relaxation. The sequence-to-vector mapping function, $F$, of our model is defined as:
\begin{equation}
    [\mathbf{x}^{(t-S+1)}, \dots, \mathbf{x}^{(t)}] \xrightarrow{F} \mathbf{\hat{y}}^{(t)}
    \label{eq:ST_eq}
\end{equation}
where $S$ is the length of the input sequence of sensor readings and each $\mathbf{x}^{(i)}\in\mathbb{R}^{D_{in}}$ is a vector of sensor measurements at timestep $i$. The model's objective is to predict the corresponding GRF vector $\mathbf{\hat{y}}^{(t)} \in \mathbb{R}^{D_{out}}$ at the final timestep $t$. For this work, the input dimension $D_{in}=8$, corresponding to the eight pressure sensors, and the output dimension $D_{out}=3$, for the force components, $[F_x, F_y, F_z]$. The input sequence length was set to $S=50$, which, given our sensor sampling rate of 1000Hz, represents a 50ms history of the contact event. To collect a diverse set of training data, a 5-axis Pocket NC V2-10 CNC machine was used. A commercial ATI Industrial Automation 6-axis force torque (F/T) sensor was mounted on the bed of the CNC machine, while the footpad sensor was mounted on the spindle. The CNC was programmed to apply a wide range of controlled trajectories such as normal, shear, and roll motions to simulate the force profiles of locomotion. We obtained $2.3 \times 10^{6}$ data points for our training dataset.

\begin{figure}
    \centering
    \includegraphics[width=0.9\linewidth]{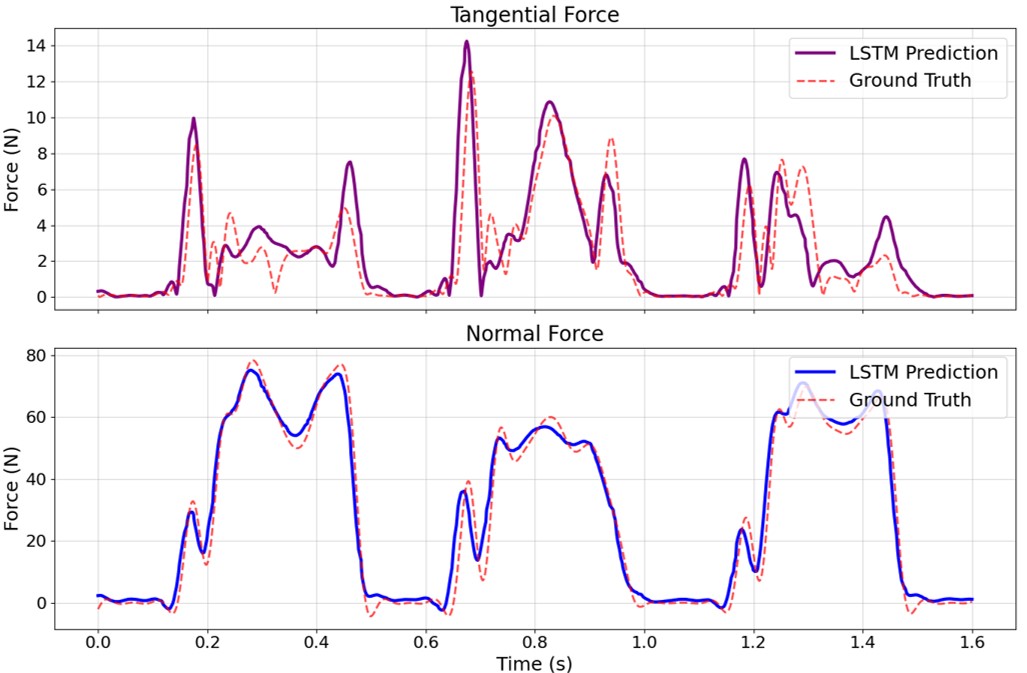}
    \caption{Comparison of simultaneous tangential and normal force predictions from LSTM model with ground truth force plate data.}
    \label{fig:lstm_force_plate}
\end{figure}

To evaluate the performance of the proposed force inference model, we integrate the sensor onto a Unitree Go1 quadruped. Fig.  \ref{fig:lstm_force_plate} compares the model predictions against ground-truth measurements from a Kistler 9260AA6 force plate during consecutive trotting motions. Our force inference latency is 0.24ms and we run the module at 500Hz. The results show good alignment between predicted and measured forces, demonstrating that the inferred forces are sufficiently accurate to serve as a reliable input for the downstream slip detection task.

\section{Anomaly Detection for Online Slip Detection}
\label{slipsense}
\begin{figure*}
    \centering
    \includegraphics[width=1.0\textwidth]{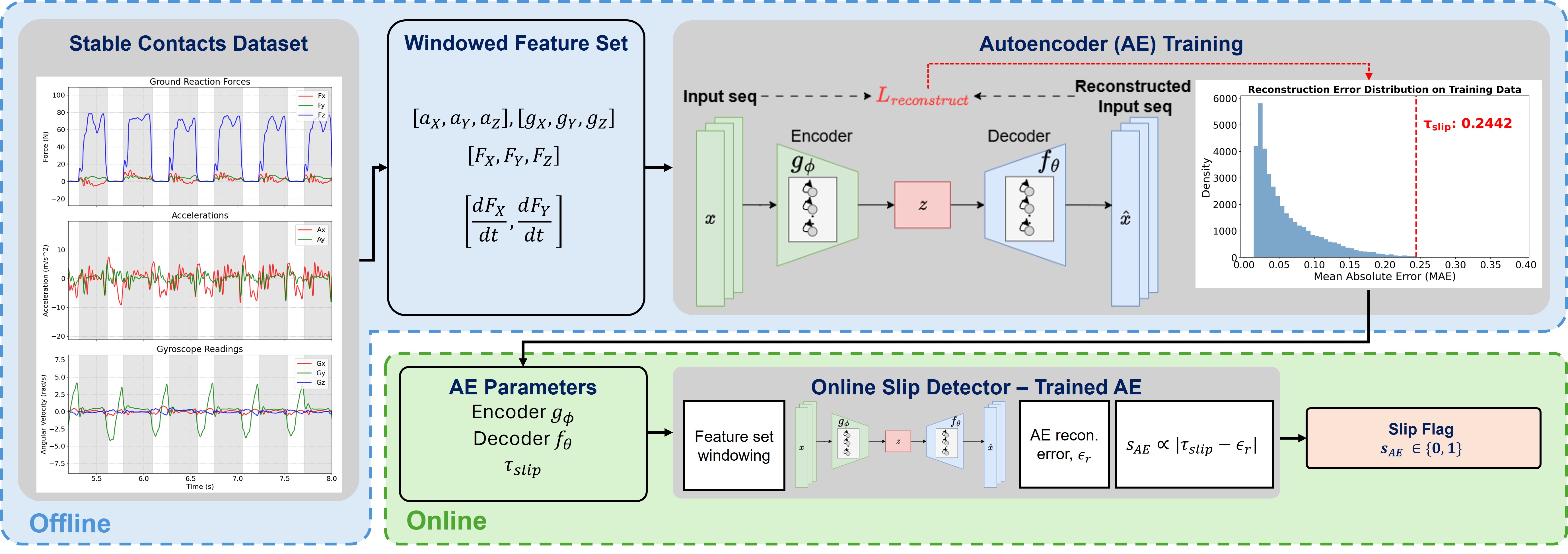}
    \caption{The proposed slip detection framework, illustrating the offline training pipeline and online deployment. This framework consists of three key modules: (1) Data collection of stable contacts and extraction of windowed features. (2) Offline self-supervised training of an LSTM AE to learn a stable contact model. The $99.5^{\text{th}}$ percentile of the reconstruction error distribution on a validation set is used to determine the threshold for slip classification, $\tau_{slip}$. (3) Online anomaly detection using the trained AE model where the real-time reconstruction error, $\epsilon_r$, is computed and compared against $\tau_{slip}$ to generate a binary slip flag, $s_{AE}$.}
    \label{fig:slip_framework}
\end{figure*}

The next component of our SlipSense framework leverages the force estimates from our force inference model to learn a robust representation of normal gait, enabling the detection of slips as anomalies from this learned behavior. This section discusses the physical principles of slip, our one-class learning methodology, and the real-time implementation. A summary of the slip detection framework is shown in Fig. \ref{fig:slip_framework}.

The interaction between the robot's foot and the ground can be defined as a stable foothold (stick) or a slip state. According to the Coulomb friction model, a stable foothold is maintained as long as the total tangential force, $F_T=\sqrt{F_x^2+F_y^2}$, exerted by the leg remains less than the maximum allowable static friction force, as described by the equation:
\begin{equation}
    F_T < \mu_s F_N
    \label{eq:stick_condition}
\end{equation}
where $\mu_s$ is the static coefficient of friction and $F_N$ is the normal force. Conversely, incipient slip occurs at the moment the stick condition is violated, ie. when $F_T \ge \mu_s F_N$, which is the point we are interested in detecting via our multimodal method. A loss of traction involves a rapid transition from static to kinetic friction, in the form of high-frequency fluctuations and large transient spikes in the shear force derivatives ($\frac{dF_X}{dt}$, $\frac{dF_Y}{dt}$) \cite{focchi2018slip}. These sharp changes in force profiles indicate an unstable contact occurring.

From a kinematics point of view, a stable foothold is defined by zero velocity of the foot relative to the ground. The foot acceleration in the world frame should therefore remain near zero, accounting only for minor vibrations and noise. We leverage kinematics data to detect fluctuations in the tangential acceleration which represent an unstable foothold. To achieve this, we utilize IMU data, which is transformed from the foot's body frame to the world frame via a Madgwick filter. Significant, non-zero accelerations in the tangential plane during stance hint towards an unstable foothold.  Our methodology, therefore, is built on the hypothesis that the onset of slip manifests as a combined anomaly in both the kinematic (inertial) and kinetic (force derivative) signals. We fuse these modalities to create a feature set for learning the characteristics of contact stability.

Acquiring a large and diverse dataset of labeled slip events is difficult and often impractical due to the dangerous and inconsistent nature of the event. To overcome this challenge, we frame slip detection as a one-class classification (or anomaly detection) problem. This approach allows us to train a model exclusively on an abundance of easily-collected data of the robot walking with stable contacts on high friction terrains. Our stable contacts dataset was collected across a variety of conditions to ensure the model learns a comprehensive representation of stability. This included multiple speeds, changes in direction, in-place rotations, and traversal over different high-traction terrains.

The core of our methodology is an LSTM Autoencoder (AE), trained via self-supervised learning on this stable contacts dataset. The training pipeline is as follows:
\begin{itemize}
    \item \textbf{Stance Phase Extraction:} By setting a normal force threshold, $F_z>5N$, we segment the time series data to isolate the stance phases.
    \item \textbf{Feature Engineering:} For each time step, a feature vector, $x_t$, is constructed and involves a fusion of kinematics (inertial data) and kinetics (shear force derivatives) information to capture the contact dynamics:
    \[x_t = \big[ a_x, a_y, a_z, g_x, g_y, g_z, \tfrac{dF_x}{dt}, \tfrac{dF_y}{dt} \big] \in \mathbb{R}^d\]
    where $d=8$ is the feature dimension. 
    \item \textbf{Windowing:} The feature vectors are segmented into moving windows of 50ms ($T=25$ samples at 500 Hz). This duration is short enough for real-time detection while being sufficiently long to capture temporal patterns within the window. Our feature window is defined as:
    \begin{equation}
        \textbf{X} = (x_1, x_2, \dots, x_T) \in \mathbb{R}^{T \times d}
    \end{equation}
    
    \item \textbf{Self-supervised Training:} The autoencoder is trained to reconstruct these feature windows. The LSTM encoder, $g_\phi$, maps the input sequence, $\textbf{X}$, into a low-dimensional latent embedding, $z \in \mathbb{R}^k$, where $k$ is the embedding dimension. We use the encoder's final hidden state, $h_T$, as the latent representation, $z = h_T \in \mathbb{R}^k$.
    
    The decoder, $f_\theta$, then attempts to reconstruct the original input sequence, $\hat{\textbf{X}} = (\hat{x}_1, \dots, \hat{x}_T)$, from the latent embedding, $z$. At each time step, the decoder uses a dense layer to output the final reconstructed feature vector:
    \begin{equation}
        \hat{x}_t = W_y h_t' + b_y
    \end{equation}
    where $h'_t \in \mathbb{R}^{k}$ is the hidden state output of the decoder LSTM at time step $t$. $W_y$ and $b_y$ are the weight matrix and the bias vector of the dense layer, respectively.
    The model is trained by minimizing the L1 reconstruction loss, which computes the mean absolute error between the original and reconstructed sequences over all time steps:
    \begin{equation}
        \mathcal{L}_{\text{rec}} = \frac{1}{T} \sum_{t=1}^T \| x_t - \hat{x}_t \|_1
    \end{equation}
    
    where $\|v\|_1 = \sum_j |v_j|$ is the L1 norm.
\end{itemize}

After training on the stable contacts dataset, the $99.5^{\text{th}}$ percentile of the reconstruction errors (as shown in Fig. \ref{fig:slip_framework}) on a validation set is then calculated and stored as the anomaly detection threshold, $\tau_{slip}$, for determining the slip label, $s_{AE}$:
\begin{equation}
    s_{AE} = 
\begin{cases}
    1, & \text{if }  \varepsilon_{r}(t) > \tau_{slip}, \\[6pt]
    0, & \text{otherwise.}
\end{cases}
\end{equation}
where $\varepsilon_{r}(t)$ is the real-time L1 reconstruction error, calculated as:
\begin{equation}
    \varepsilon_{r}(t) = \frac{1}{T} \sum_{i=1}^{T} || \mathbf{x}_{t-T+i} - \hat{\mathbf{x}}_{t-T+i} ||_1
\end{equation}

\section{Experimental Validation of SlipSense on a Quadruped Robot}
The trained AE model is deployed to evaluate the performance of slip detection on a real quadruped robot. This section introduces the experimental setup, the ground truth labeling methodology, the baseline model, and the experimental results.
\subsection{Experimental Setup}
\begin{figure}
    \centering
    \includegraphics[width=1.0\linewidth]{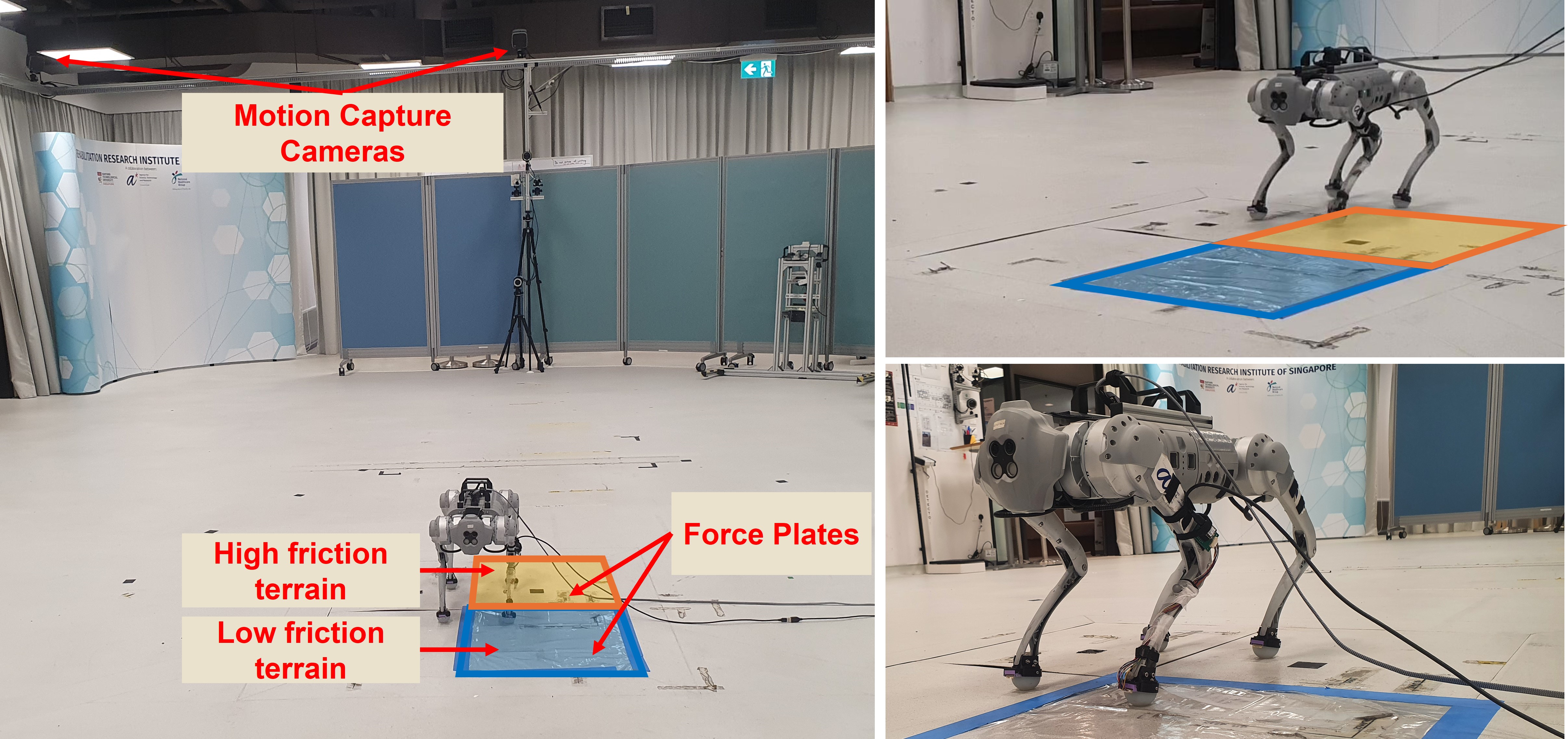}
    \caption{Experimental setup in motion capture lab with force plates. The setup involves a setup with two force plates with a non-slippery terrain (orange) and a slippery terrain (blue).}
    \label{fig:force_plate_setup}
\end{figure}

The experimental validation was conducted in a motion capture lab equipped with a Qualisys system running at 200Hz, using a mix of Miqus M3 and Arqus A12 cameras (15 cameras in total) with resolutions of 2MP and 12MP, respectively. The floor was integrated with two Kistler 9260AA6 force plates sampling at 2kHz, providing ground-truth GRFs. The setup is shown in Fig. \ref{fig:force_plate_setup}. The sensor was mounted on all four feet of the Unitree Go1 quadruped. However, for clarity of evaluation and to isolate the sensing and inference pipeline, only the front-left foot was used for real-time force inference and slip detection in this study. The robot was tasked with trotting across transitioning terrains consisting of a high-friction laboratory floor ($\mu_s \approx 0.8$) and a low-friction patch created using a vinyl sheet sprayed with WD40 lubricant ($\mu_s \approx 0.1$) with speeds up to $0.3m/s$. Three reflective markers were placed on the front-left leg (one on the foot and two on the shank forming a triangular plane) to track the relative orientation between the sensor frame and the force plate frame. The marker directly on the foot was used for extracting foot positions. During trials, we recorded motion capture kinematics, force plate GRFs, force sensor readings, and slip detection outputs from both our method and a kinematic baseline.

\subsection{Ground Truth Labeling}
\begin{figure*}
    \centering
    \includegraphics[width=1.0\textwidth]{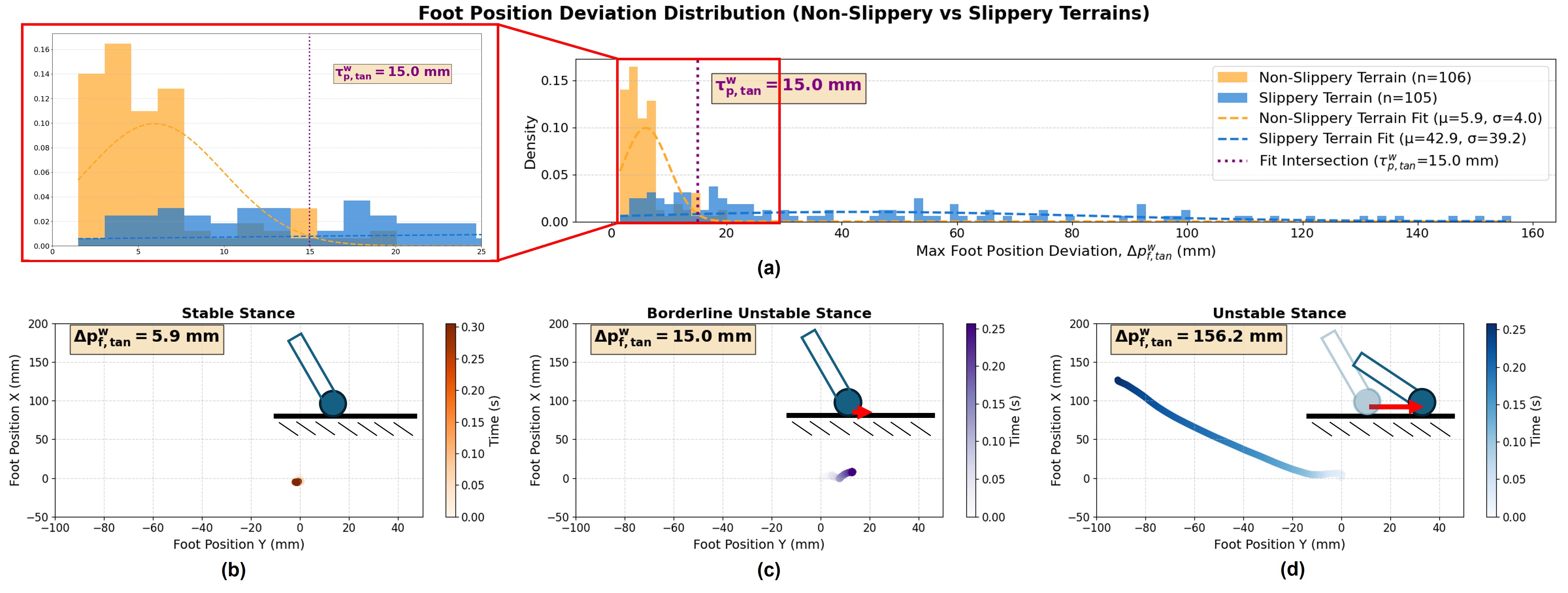}
    \caption{Data-driven ground truth labeling and visualization of different slips. (a) distributions of foot position deviations, $\Delta p_{f,tan}^w$, using motion capture for non-slippery (orange) and slippery (blue) terrains and the determined threshold, $\tau_{p,tan}^w$. Single stance foot positions over time are shown in (b) stable foothold ($\Delta p_{f,tan}^w<\tau_{p,tan}^w$), (c) borderline slip stance ($\Delta p_{f,tan}^w = \tau_{p,tan}^w$), and (d) extreme slip case ($\Delta p_{f,tan}^w>\tau_{p,tan}^w$).}
    \label{fig:COP}
\end{figure*}
To obtain ground truth slip labels, we leveraged the motion capture system to track the tangential foot position during stance. Note that in similar works such as \cite{yan2024slip}, the ground truth labels were simply determined based on whether the foot is on the slippery patch or not. However, this may result in false labeling as there are cases where slipping occurs on the non-slippery patch due to residue, or no slipping on the slippery patch. In our work, a stance is labeled as slip if the deviation of the foot position from its initial contact point, $\Delta p_{f,\text{tan}}^{w}$, exceeds a threshold, as calculated from:
\begin{equation}
\Delta p_{f,\text{tan}}^{w} = 
\sqrt{\big(x_{f}^{w}(t) - x_{f}^{w}(t_{0})\big)^{2} 
+ \big(y_{f}^{w}(t) - y_{f}^{w}(t_{0})\big)^{2}},
\end{equation}

where $t_{0}$ denotes the foot touchdown. The ground truth slip label, $s$, is therefore defined as:
\begin{equation}
s =
\begin{cases}
1, & \text{if } \Delta p_{f,\text{tan}}^{w} > \tau_{p,\text{tan}}^{w}, \\
0, & \text{otherwise}.
\end{cases}
\end{equation}
 
 To determine the threshold, $\tau_{p,\text{tan}}^{w}$, we adopt a data-driven approach (Fig. \ref{fig:COP}(a)). We collected foot position data over 211 stances across both non-slippery (orange) and slippery terrains (blue). For each stance, we computed $\Delta p_{f,\text{tan}}^{w}$. The resulting distributions are plotted and fitted with Gaussian curves, and the intersection point between the two distributions is chosen as the threshold for ground truth labeling. Fig. \ref{fig:COP}(b)-(d) further illustrates example stance trajectories, showing a stable stance, a borderline unstable stance, and an unstable stance with over 150mm of foot displacement. Given the difficulty in explicitly labeling slip, this approach provides a robust, data-driven method for generating reliable slip labels to validate our framework.

\subsection{Baseline Slip Detection Model}
\label{chap:baseline}
To assess the performance of our slip detection framework, we implemented a baseline model using a common proprioceptive method based on kinematics. The core principle, discussed in works such as \cite{focchi2018slip}, is that a stable foothold should exhibit near-zero velocity.  This world-frame velocity check serves as a point of comparison and its components are integrated in various proprioceptive detectors \cite{yan2024slip}\cite{nistico2022slip}. To ensure a rigorous and fair comparison, the baseline's velocity threshold was determined using the same data-driven, one-class methodology and stable contact dataset used for our SlipSense framework. The foot velocity in the world frame, $\mathbf{v}_{f}^{w}$, is computed as
\begin{equation}
    \mathbf{v}_{f}^{w} = \mathbf{v}_{b}^{w} + \mathbf{R}_{b}^{w} \big( \mathbf{v}_{f}^{b} + \boldsymbol{\omega}^{b} \times \mathbf{p}_{f}^{b} \big)
    \label{eq:foot_velocity}
\end{equation}
where $\mathbf{v}_{b}^{w}$ is the robot base linear velocity in world frame, $\mathbf{R}_{b}^w$ is the rotation matrix from base to world frame obtained from the base IMU quaternion, $\mathbf{v}_{f}^{b}$ is the foot velocity expressed in the base frame, $\boldsymbol{\omega}_{b}^{b}$ is the base angular velocity, and $\mathbf{p}_{f}^{b}$ is the foot position in the base frame. The tangential component is then extracted as
\begin{equation}
    v_{f,\text{tan}}^{w} = \sqrt{ \big(v_{f,x}^{w}\big)^{2} + \big(v_{f,y}^{w}\big)^{2} }.
\end{equation}

We extract $v_{f,\text{tan}}^{w}$ values across stance phases and set the $99.5^{\text{th}}$ percentile as the slip threshold, $\tau_{vel}$. During inference, slip is flagged whenever $v_{f,\text{tan}}^{w}$ exceeds this threshold, as represented by:
\begin{equation}
    s_{baseline} = 
\begin{cases}
    1, & \text{if } v_{f,\text{tan}}^{w} > \tau_{vel}, \\[6pt]
    0, & \text{otherwise.}
\end{cases}
\end{equation}

\subsection{Results: Online Slip Detection}

\begin{figure*}
    \centering
    \includegraphics[width=1.0\textwidth]{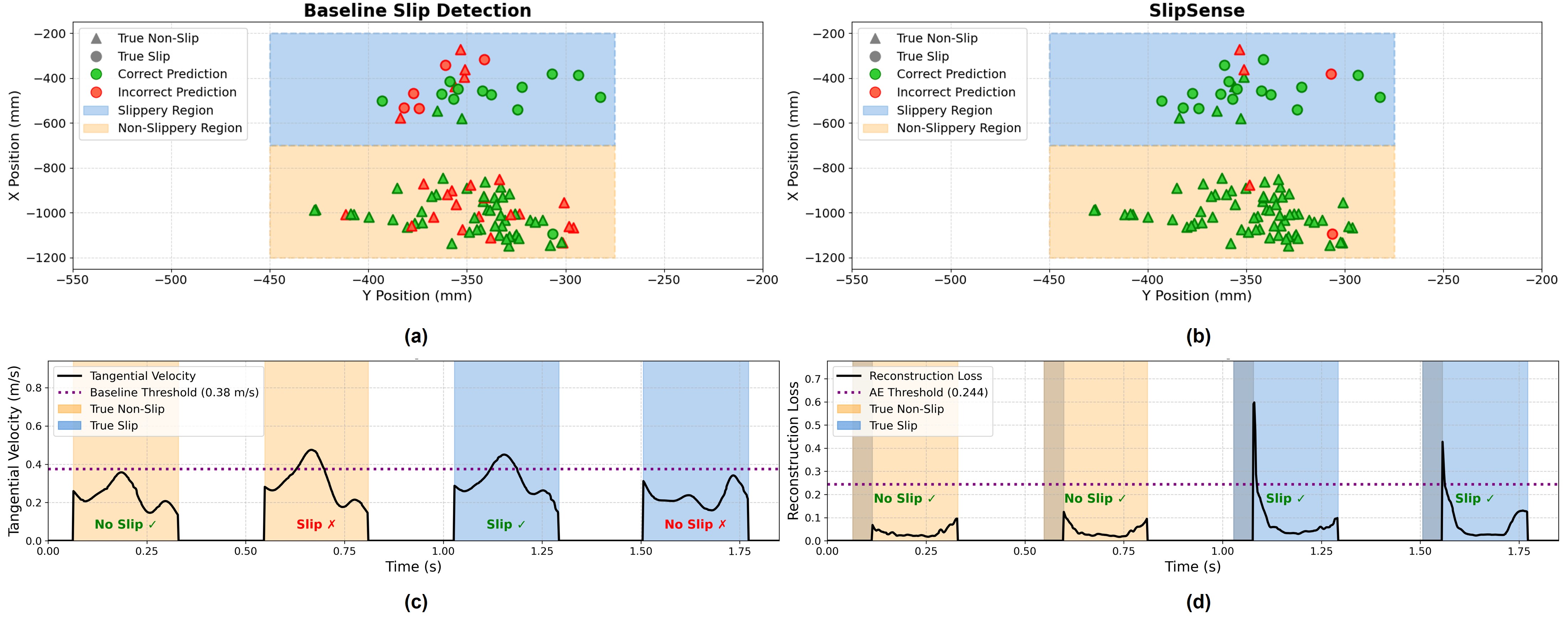}
    \caption{Experimental results comparison between baseline and the proposed multimodal SlipSense slip detection models. Shown are 100 stances and the corresponding correct/incorrect predictions for (a) baseline model and (b) SlipSense. The time series plot over four stances are shown and the corresponding thresholds plotted for the (c) baseline and (d) SlipSense models.}
    \label{fig:birdseye}
\end{figure*}

We experimentally validate the performance of our SlipSense framework against the baseline. For over 200 stance sequences, we observe an accuracy of 85.9\% on the multimodal SlipSense feature set which significantly outperforms the kinematics baseline's 69.3\%, as shown in Table \ref{tab:performance_simple}. 
An ablation study was performed to provide insight into the necessity of multimodal fusion. The AE model was trained using two feature set variants -- force-only and inertial-only. The force-only model, while more accurate than the baseline, struggled to differentiate between slip-induced force behavior and intended initial ground impact effects, limiting its accuracy to 73.7\%. Conversely, the inertial-only model achieved a higher accuracy of 83.9\%, demonstrating the benefit of gathering inertial information right at the end effector, but its sensitivity was limited. By capturing the high-frequency transient spikes as foot slippage occurs, the force derivatives enable the detection of early slips, justifying the multimodal fusion.

\begin{table}[t]
\centering
\caption{Slip detection performance and resolution across models.}
\label{tab:performance_simple}
\renewcommand{\arraystretch}{1.15}
\setlength{\tabcolsep}{3.5pt} 
\begin{tabular}{lccc}
\toprule
\textbf{Model} & 
\makecell{\textbf{Accuracy} \\ (\%)} & 
\makecell{\textbf{Smallest Slips$^1$} \\ (mm)} & 
\makecell{\textbf{Min. Detected} \\ \textbf{Slip} (mm)} \\
\midrule
Baseline Detector & 69.3 & 80.8 $\pm$ 35.6 & 16.5 \\
SlipSense (Force-only) & 73.7 & \textbf{20.7 $\pm$ 4.2} & \textbf{15.4} \\
SlipSense (IMU-only) & 83.9 & 37.3 $\pm$ 15.5 & 15.6 \\
SlipSense (Multimodal) & \textbf{85.9} & 24.1 $\pm$ 6.4 & \textbf{15.4} \\
\bottomrule
\multicolumn{4}{l}{\footnotesize $^1$Mean $\pm$ Std. Dev. over the 30 smallest correctly classified slips.}
\end{tabular}
\end{table}

Fig. \ref{fig:birdseye}(a)-(b) shows 100 stance phases of the robot traversing over two terrains of different friction to qualitatively evaluate the performance of the two models. The baseline model exhibits two primary failure modes. Firstly, a significant number of false positives on the non-slippery terrain is observed, where stable stances are incorrectly flagged as slip. This is attributed to noise in the world frame foot velocity estimate and the inability of the model to differentiate between a slip and intended ground impact. Secondly, a large proportion of false negatives is observed, in particular for low-displacement slips that occur near the ground truth 15mm boundary. In contrast, our SlipSense demonstrates superior performance over the baseline, with a significant reduction in false positives and negatives. A comparison of the time-series plots in Fig. \ref{fig:birdseye}(c)-(d) shows our model's reconstruction error remains low during non-slip stances and spikes under the slip conditions. Over four consecutive stances, our model successfully classified the slip state, while the baseline model resulted in a false positive and false negative.

\begin{figure}
    \centering
    \includegraphics[width=1.0\linewidth]{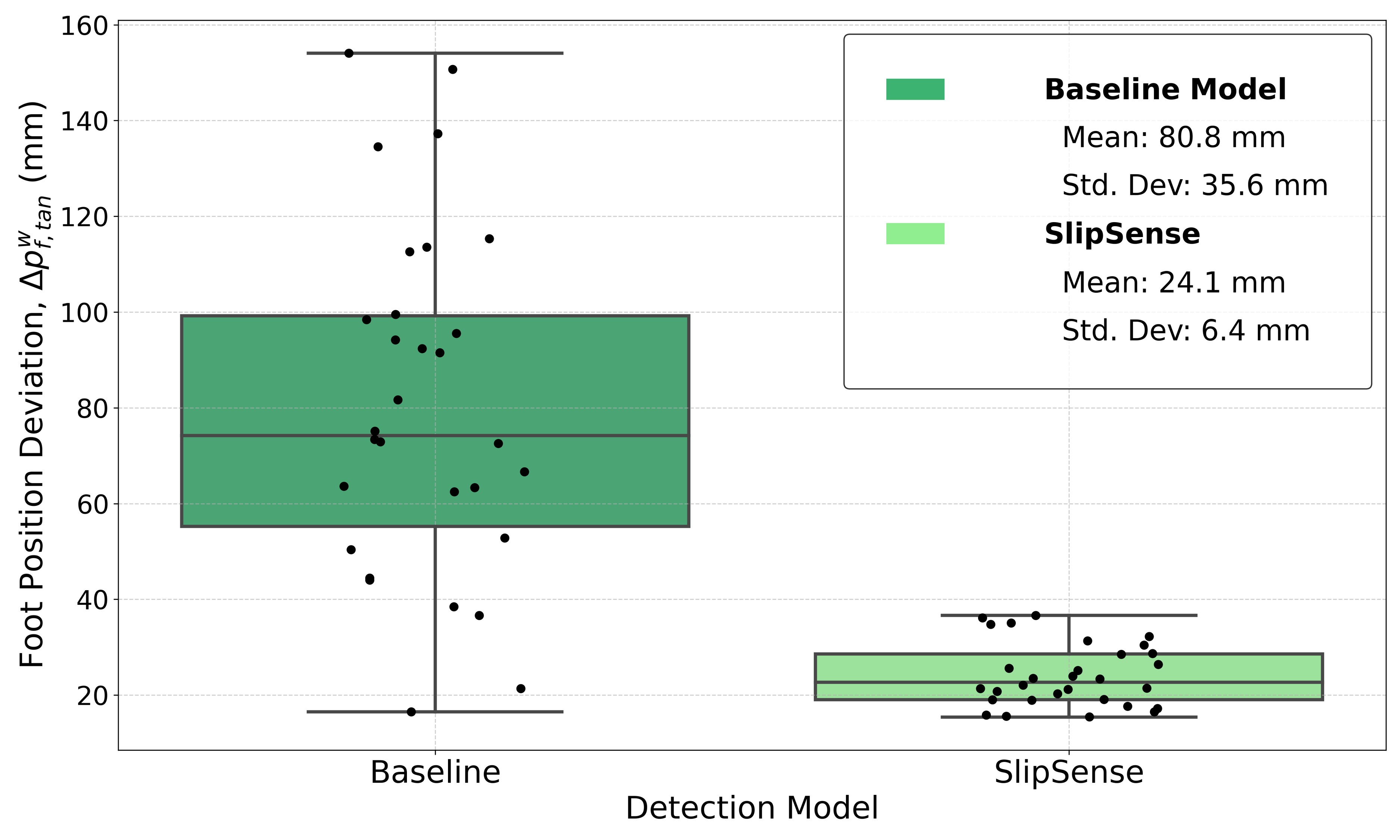}
    \caption{Box plot showing the smallest 30 slip stances accurately classified for the baseline and SlipSense models.}
    \label{fig:boxplot}
\end{figure}

A key contribution of our work is the improved resolution of slip detection, as shown in Table \ref{tab:performance_simple} and Fig. \ref{fig:boxplot}. Over the 30 smallest correctly detected slip stances, which represents the best case scenario for each model, the baseline model was only able to detect slips with an average minimum displacement of 80.8 ± 35.6mm, which represents significant gross slips and large variability. Our multimodal SlipSense model, however, achieved an average minimum detected slip of 24.1 ± 6.4mm; a 3.3-fold improvement from baseline. The significantly lower mean and standard deviation demonstrates its superior ability to detect smaller slips more consistently. Detection of slips with 24.1mm foot displacement provides a crucial early warning to the downstream controller, enabling ample time for recovery and mitigation strategies before catastrophic instability.

\section{Conclusion and Future Work}
This paper presents SlipSense, an end-to-end framework for online slip detection using a custom multimodal force sensor. By fusing kinetic and kinematic data within a self-supervised anomaly detection model, our framework demonstrates a 3.3-fold improvement in slip detection resolution and higher overall accuracy over a velocity-level kinematic baseline.

While promising for integration into terrain-aware controllers, we acknowledge limitations that provide clear avenues for future research. The model was validated on rigid, homogeneous surfaces with a set trotting gait. Extension to unstructured ground or inclines requires improved generalization in the one-class learning. Secondly, our current slip detection involves a binary classification task in which a data-driven labeling method is adopted, which introduces ambiguity near the threshold. The AE model shows significantly more errors near this threshold. Thus, we look to create a third class of incipient slip near the boundary to better represent the labeling function.

Future work will involve optimizing the force inference model to improve the tangential force estimates which will improve the downstream slip detection task overall. SlipSense will be extended to perform online terrain friction estimation which can be integrated into a robot controller to perform slip recovery and terrain-aware gait adaptation. Finally, integration of all four foot force sensors will also be explored. The overall paper forms the foundation for future force-based gait adaptation control.

\section*{Acknowledgments}
Iris Szu-Yao Liu is an A*STAR scholar under the Singapore International Graduate Award (SINGA) framework. The authors would like to thank Yuehang (Aria) She and Alex Lim Lek Syn for assisting with the experiments. 

\bibliographystyle{unsrt}
\bibliography{bib}

\vspace{11pt}
\vfill
\end{document}